# Ranking Pages by Topology and Popularity within Web Sites

José Borges
School of Engineering,
University of Porto
R. Dr. Roberto Frias, 4200 - Porto, Portugal
`jlborges@fe.up.pt`

Mark Levene
School of Computer Science and Information Systems,
Birkbeck, University of London
Malet Street, London WC1E 7HX, U.K.
`mark@dcs.bbk.ac.uk`

**Abstract**

We compare two link analysis ranking methods of web pages in a site. The first, called *Site Rank*, is an adaptation of PageRank to the granularity of a web site and the second, called *Popularity Rank*, is based on the frequencies of user clicks on the outlinks in a page that are captured by navigation sessions of users through the web site. We ran experiments on artificially created web sites of different sizes and on two real data sets, employing the relative entropy to compare the distributions of the two ranking methods. For the real data sets we also employ a nonparametric measure, called Spearman's footrule, which we use to compare the top-ten web pages ranked by the two methods. Our main result is that the distributions of the Popularity Rank and Site Rank are surprisingly close to each other, implying that the topology of a web site is very instrumental in guiding users through the site. Thus, in practice, the Site Rank provides a reasonable first order approximation of the aggregate behaviour of users within a web site given by the Popularity Rank.

## 1 Introduction

Link analysis for ranking web pages [Hen01] has been a very active subarea in the past few years. The most influential techniques have been HITS (Hyperlink-Induced Topic Search) [Kle99], which computes hub and authority scores for web pages related to a search query, and Google's PageRank [BP98, PBMW98], which computes a query independent ranking of web pages. Due to the 'high profile' of Google and the simple yet expressive formulation of PageRank in terms of a Markov chain [KS60], there have been several studies of its fundamental properties [BGS04, LM04]. Herein we will only consider link analysis with respect to a web site rather than for the whole web as a global search engine such as Google does.

Data mining of user navigation patterns is an important activity within web usage mining [Mob04], whose aim is to build a model of the users surfing a web site from server log data over a time period. The method we have been using for this purpose builds a Markov chain that represents user navigation sessions from which further data mining, such as finding the frequent trails that were followed, can be deployed [BL00a, BL00b]. The entropy [Khi57] of the induced Markov chain is a measure of the uncertainty of navigating through the underlying web site; if the entropy is low the trails are more predictable while if the entropy if high they are more 'random'. In [LL03] we presented an algorithm for computing the entropy of the chain via a random walk on the Markov chain that also returns its stationary distribution, assuming that the chain is ergodic [KS60].

We utilise and extend the algorithm developed in [LL03] to compute two rankings of web pages within a web site. The first is *Site Rank*, which can be viewed as an adaptation of PageRank to the granularity of a web site, and the second is *Popularity Rank*, which is based on the navigation sessions of users through the web site. Both Site Rank and Popularity Rank can be computed via a random walk on the web site. For Site Rank we assume that transitions from one page to another are random uniform, while for Popularity Rank we assume that the transitions are taken according to the frequencies of user clicks on the outlinks in a page. The computation of Popularity Rank is highly efficient as it can be carried out by a single scan of the input data, which we assume is pre-processed into navigation sessions. The computation of Site Rank is also efficient but depends on the length of the random walk on the Markov chain, necessary for convergence to the stationary distribution. We also present an extension of the Popularity Rank to include links to pages that were not visited at all, and thus have not record in the server log. This is useful to webmasters, as it may give them insight into the reasons behind the unpopularity of certain web pages.

The primary aim of this work is to compare Popularity Rank with Site Rank. To achieve this we ran experiments on artificially created web sites of different sizes, and on two real data sets. To assess how close the distributions of the Popularity Rank and Site Rank are, we employ the relative entropy [Gra90] and provide a novel normalisation of it using the notion of the maximum relative entropy. In addition, for the real data sets we employ a nonparametric measure, called Spearman's footrule [FKS03], which we use to compare the top-ten web pages ranked by the two methods. Our main result is that the distributions of the Popularity Rank and Site Rank are surprisingly close to each other, implying that the topology of a web site is very instrumental in guiding users through the site. Of course it is possible that a page with low Site Rank will be highly popular or vice versa, but, overall, it seems that this is *not* the case. From the empirical evidence we have gathered and by simulation on artificially created web sites, it appears that, in practice, Site Rank is a reasonable first order approximation of the aggregate behaviour of users within a web site as given by the Popularity Rank.

The rest of the paper is organised as follows. In Section 2 we introduce the Popularity Rank and Site Rank through an example web site and illustrate the methods we use for computing the entropy of a web site, and the relative entropy between the Site Rank and Popularity Rank. In Section 3 we present the results of experiments with the page ranking methods on synthetic data sets (Subsection 3.1) and on two real data sets (Subsection 3.2). Finally, in Section 4 we give our concluding remarks.

## 2  Popularity Rank and Site Rank

When describing users' activity within a web site we consider the *navigation sessions* of users browsing pages on the site, where a navigation session is a sequence of page requests from the site for a single user. We note that there are several heuristics for reconstructing navigation sessions from web server logs [SMBN03], the most common one involving user identification by IP address and restricting the total duration of the session so that it be no longer than 30 minutes.

Both the PageRank and the Popularity Rank of a web site can be defined in terms of the stationary distribution of an ergodic finite Markov chain [KS60]; by ergodic we mean here both irreducible and aperiodic. To demonstrate this we view the web pages of a site as the states of a Markov chain and impose transition probabilities on the links of each page. We assume that the web site has a distinguished web page, call it the *home page*, such that all navigation sessions start and finish from the home page [LL03]. (To ensure that the chain is ergodic we may add a link from the home page to itself.) This technique of attaining ergodicity is similar to the method used in [LM04] that adds an artificial web page to the site that links to all other pages and is also linked to by all pages. The technical difference between these methods is that, in our method we do not insist that the home page is a source and sink for *all* other pages, since in our model all web pages are reachable from the home page.

The PageRank is based on the 'random surfer' model, where the user, when confronted with the set of links embedded on the page being browsed, chooses uniformly at random the link that will be clicked on next (see [PBMW98]). So, in this case, the resulting Markov chain is one where, from each state representing a web page, the transition probabilities are divided equally between each of the outlinks. We call this variation PageRank restricted to a single web site *Site Rank*.

On the other hand, the Popularity Rank is based on a model, where the transition probabilities of outlinks are determined by usage according to server web logs, that is, by the number of times the link was previously clicked on [LL03].

In [LL03] we describe a method to compute the entropy of an ergodic Markov chain via a random walk on the chain that also returns its stationary distribution. We note that the stationary distribution can also be computed using matrix computation methods such as the ones described in [BGS04, LM04]. In the case of computing the Popularity Rank from web server logs, our algorithm is linear time in the combined size of the navigation sessions. To see this assume a fixed web site that has $N$ pages and $L$ links, and that $t$ is the sums of the lengths of the navigation sessions. Let

$$\pi = \left\{ \frac{m_i}{t} \right\}_i \quad \text{and} \quad P = \left\{ \frac{m_{ij}}{m_i} = P_{ij} \right\}_{ij}, \tag{1}$$

be, respectively, the proportion of time page $i$ was visited and the proportion the page $j$ was visited after page $i$, where $i, j \in \{1, \ldots, n\}$. Then it is easily verified that $\pi$ is the stationary distribution of the chain, since

$$\pi = \left\{ \frac{m_i}{t} \right\}_i = \left\{ \sum_j \frac{m_i}{t} \frac{m_{ij}}{m_i} \right\}_i = \pi P. \tag{2}$$

Letting $P$ denote the Popularity Rank distribution and $Q$ denote the Site Rank distribution, whose probabilities $\{Q_{ij}\}$ are defined according to the random surfer model described above. Then the *relative entropy* between the Popularity Rank distribution (according to a given server log) and the Site Rank distribution is defined as

$$D(P\|Q) = \sum_{i,j} P_{ij} \log \frac{P_{ij}}{Q_{ij}}, \tag{3}$$

the interpretation in our context being a measure of how close the Popularity Rank is to the Site Rank. (As usual logarithms are taken to the base 2, $D(P\|Q) \geq 0$ and if $P = Q$ then $D(P\|Q) = 0$ [Gra90].)

In the extreme case, where for each page $i$ there is a page $j$ such that $P_{ij} = 1$

$$D^{max}(P\|Q) = \sum_i \log(d_i), \tag{4}$$

where $d_i$ is the out-degree of page $i$. We note that this corresponds to the situation when the server log contains only a single trail induced by the navigation sessions, where the user browsing a web page always choose the same link each time he or she visited the page. It can easily be shown that, assuming $Q$ is fixed to be the Site Rank distribution, (4) is the maximum relative entropy between $P$ and $Q$ and thus can be used to normalise the relative entropy.

One problem with the Popularity Rank as we have defined it is that some web pages in the site may not have been visited at all, and thus will not be recorded in the server log. It is important for web masters to have this information and understand the reason behind the unpopularity of these unvisited pages, in order to make a decision whether any remedial action needs to be taken.

This motivates an extension of the Popularity Rank as defined in [LL03] as follows. We call a link in the web site *popular* if it has been traversed at least once, i.e. the link appears in at least one navigation session being inferred from the server log, and a link which has not been traversed an *unpopular* link. Letting $u_i$ be the number of unpopular links outgoing from page $i$, we redefine the transition probability from page $i$ to page $j$ as

$$\frac{m_{ij}}{m_i + u_i} \quad \text{and} \quad \frac{1}{m_i + u_i}, \tag{5}$$

when the link from $i$ to $j$ is popular and, respectively, unpopular. We denote the modification of the distribution of the Popularity Rank in this manner by $P'$.

To compute the Popularity Rank using this modification that takes unpopular links into account we invoke the random walk method, described above, using $P'$ as input to the algorithm. We note that while computing the Popularity Rank from $P$ does not require apriori knowledge of the topology of the web site, computing it from $P'$ requires such knowledge. (Also, note that computing the Site Rank from $Q$ requires apriori knowledge of the topology of the web site.)

| Session | NOS |
|---|---|
| HP,$A_1, A_2, A_3$,HP | 3 |
| HP,$A_4, A_1, A_4, A_2, A_3$,HP | 2 |
| HP,$A_1, A_4, A_3$,HP | 2 |
| HP,$A_1, A_4, A_2$,HP | 4 |

Table 1: An example of a collection of user web navigation sessions

Consider the collection of user navigation sessions given in Table 1, where NOS represents the number of occurrences of each session. Each navigation session begins and terminates at the home page of the site, which we denote as HP. From the collection of sessions a first-order Markov model can be incrementally inferred; the final model inferred from the example is represented in Figure 1. Next to a transition in the figure, the first number indicates the number of times the corresponding link was traversed and the number in parentheses represents the estimated transition probability. In order to force the chain to be aperiodic, in addition to being irreducible, we add a loop to HP. While building the model each navigation session can be viewed as a walk through the site, and the method given in [LL03] to compute the entropy of an ergodic Markov chain can be applied. The application of the algorithm given in [LL03] gives $H = 44.42$, $t = 49$ and therefore $H/t = 0.91$, where $t$ is the length of the walk and $H/t$ is the per-step entropy, which is an incremental version of the theoretical entropy given by (6) below, updated at every step of the random walk.

In addition, the stationary distribution, $\pi$, obtained from user data is

$$\pi = \{0.25, 0.23, 0.18, 0.14, 0.20\}.$$

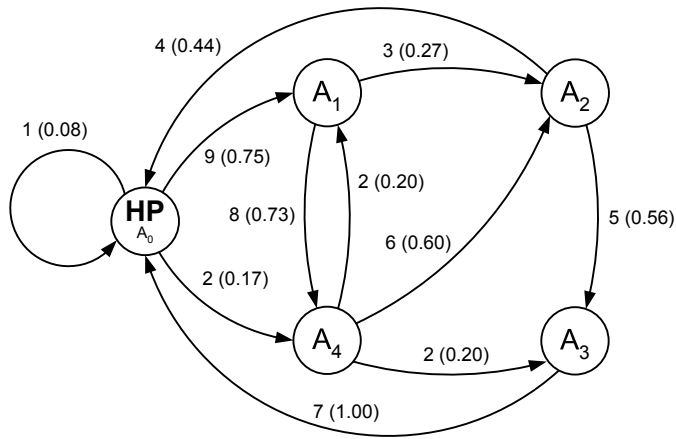

Figure 1: Model inferred from user data

The accuracy of the entropy estimate $H/t$, obtained from user data can be assessed by the theoretical expression for the entropy of a Markov chain [Khi57], given by

$$H(theory) = -\sum_{i=0}^{4}\sum_{j=0}^{4} \pi_i P_{ij} \log P_{ij} = 0.91. \tag{6}$$

The Site Rank, can be estimated by a random walk. To carry out the computation we set all transition probabilities according to a uniform distribution as follows:

$$P_{0,i} = 1/N \text{ and } P_{ij} = 1/d_i,$$

where $P_{0,i}$ stands for transitions from the HP, $N$ for the number of pages in the site ($N = 5$ in the example) and $d_i$ for the out-degree of page $i$. For the example, a random walk was performed with $t = 180$ and the estimate obtained for the entropy was $H = 242.44$, therefore, $H/t = 1.35$. (We have set the length of the random walk to be $t = 10 \cdot (N + \sum_i d_i)$, that is, the random walk terminates the next time HP is reached after the specified $t$ is obtained.)

To assess the quality of the Site Rank estimate, we compute the standard PageRank for the site by iteratively applying the PageRank formula with a teleportation probability of zero. The application of this method to the running example gives the result

$$\pi = \{0.37, 0.12, 0.18, 0.20, 0.13\},$$

and, therefore, $H(theory)$ for the Site Rank is:

$$H(theory) = -(0.37 \log(5) + 0.12 \log(2) + 0.18 \log(2) + 0.20 \log(1) + 0.13 \log(3)) = 1.36.$$

Finally, we compute the relative entropy between the probabilities given by the user navigation sessions (i.e., the Popularity Rank distribution, $P$) and the corresponding probabilities given by the random walk (i.e., the Site Rank distribution, $Q$), using (3), as

$$D(P\|Q) = 1.66,$$

and the maximum relative entropy according to (4) is

$$D^{max}(P\|Q) = \log 5 + \log 2 + \log 2 + \log 1 + \log 3 = 5.91.$$

Now, if we have access to the topology of the web site, we can take unpopular links into account in the computations. For example, suppose that in addition to the transitions with nonzero probability as shown in Figure 1, we have the three links, $(2, 1), (3, 2)$ and $(3, 4)$, that were present in the topology of the site but were not traversed by a user in any session. When considering the site's topology, transitions to HP that were not traversed are also taken to be unpopular links, and a transition from HP to $A_3$ is added to balance the number of incoming and outgoing transitions to the page. Figure 2 shows the modified model with unpopular links included, where unpopular links are represented by dashed lines.

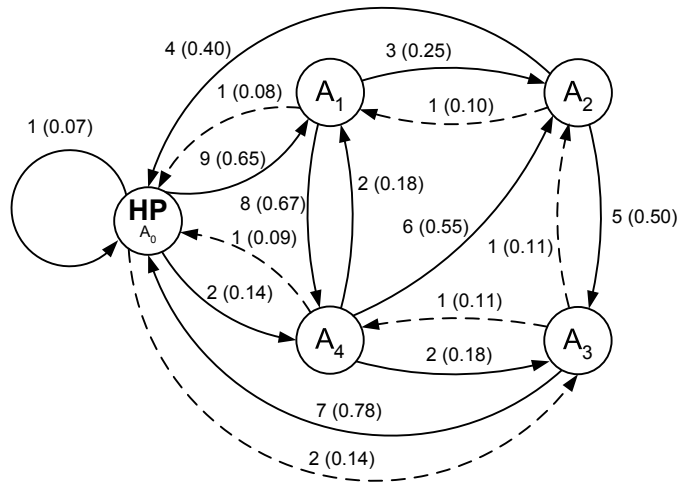

Figure 2: Model modified with unpopular links

Given the new stationary vector, which takes into account unpopular links, that is

$$\pi = \{0.25, 0.21, 0.18, 0.16, 0.20\},$$

the new estimate for $H(theory)$ for Popularity Rank is

$$H(theory) = 1.36.$$

Again, an estimate for the Popularity Rank can be obtained by a random walk on the model represented in Figure 2. A random walk with length $t = 280$ gives $H = 411.849$ and therefore $H/t = 1.47$.

The rest of the computations are similar to the case without unpopular links. A random walk (with random uniform transition probabilities) of length $t = 280$ gives an entropy estimate $H = 517.40$ for Site Rank and therefore $H/t = 1.85$. The stationary distribution for Site Rank with unpopular links included is

$$\pi = \{0.28, 0.17, 0.21, 0.17, 0.17\},$$

and

$$H(theory) = -(0.28 \log(5) + 0.17 \log(3) + 0.21 \log(3) + 0.17 \log(3) + 0.17 \log(4)) = 1.86.$$

Finally, the relative entropy for the case including unpopular links is

$$D(P\|Q) = 2.37$$

and the maximum relative entropy is given by

$$D^{max}(P\|Q) = \log 5 + \log 3 + \log 3 + \log 3 + \log 4 = 9.08.$$

Normalising the relative entropy, we get $1.66/5.91 = 0.2809$ without unpopular links, and $2.37/9.08 = 0.2610$ with unpopular links, which in this case indicates that the closeness of the Popularity Rank to Site Rank is not much affected by the addition of unpopular links.

## 3 Experiments

In order to assess the properties of the proposed ranking methods in a real-world scenario and also in a set of a configurable wider variety of possible scenarios, we conducted a set of experiments on both real and artificial data sets.

Our method for generating artificial data is divided into two stages: (i) creating a web topology, and (ii) creating a collection of user sessions based on the generated topology.

The first stage is configurable by three parameters: the number of pages in the site, $N$, an exponent for a power law to describe the inlink distribution and an exponent for a power law to describe the outlink distribution [BKM+00]. The method works as follows. For each of the $N$ pages we generate its number of outlinks and inlinks according to the corresponding power-law distributions. Then, links between pages are created by randomly matching inlink and outlink references.

The second stage makes use of the PageRank [BP98] as a measure of the user's interest in each page within the web site. In fact, the initial probability of a page is estimated by its PageRank, and the transition probabilities are estimated according to PageRank's random surfer model. The number of sessions is a parameter and the length of a session is set according to a power-law distribution [HPPL98]. For each session, we randomly choose the start page according to the initial probabilities and thereafter the next link to follow is chosen according to the probabilities of the outlinks from the current page. Also, as in the random surfer model, for each page there is a probability of 0.15 for the session terminating.

In the real data experiments we use two different data sets. The first is from a university site that was made available by the authors of [BMSW01]. This data is based on a random sample of two weeks of usage data from a university site during April 2002. According to the authors, during the time of collection caching was prohibited by the site, and the site was cookie-based in order to ease user identification. The data was made available with the sessions already identified; according to [BMSW01] sessions were inferred based on the cookie information. The second data set was obtained from the authors of [PE00] and corresponds to ten days of usage records from the site http://machines.hyperreal.org (Music-Machines) in 1999. The data made available was organised into sessions and, according to the authors, caching was disabled during the collection.

## 3.1 Experiments on Synthetic Data

Topologies of varying sizes and corresponding collections of navigation sessions were generated; the number of navigation sessions was set as $1.2 \cdot N$, which is proportional to the number of pages in the web site. The Popularity Rank was estimated by a random walk on the model inferred from the collection of sessions and the Site Rank estimate was obtained via a random walk over the topology, while taking into account unpopular links. For both random walks, the duration (measured as the number of links to follow) was set as $10 \cdot (N + L)$, in order to ensure that a large proportion of the links are traversed during the random walk.

Figure 3 shows how the entropy estimate, $H/t$, of the induced Markov chain [LL03], varies with the number of pages in the web topology, where $t$ is the length of the random walk over the topology. For comparison purposes the figure also shows the theoretical value, $H$(theory), of the entropy of the induced Markov chain as given (6). It can be seen that $H/t$ *underestimates* $H$(theory) both for Popularity Rank and Site Rank as we expect from the results in [LL03]. Moreover, the results show that the Site Rank estimate is better than the Popularity Rank estimate.

Figure 4 shows how the relative entropy between the Popularity Rank and Site Rank varies with the number of pages in the web topology, where the relative entropy, given in (3), is normalised by its maximum value given in (4). The variation of the maximum value of the relative entropy with the number of pages in the web site is shown in Table 2. It can be seen that the normalised relative entropy is more or less independent of the size of the web site and is very small, indicating that the Popularity Rank and Site Rank distributions are very close.

## 3.2 Experiments on Real Data

Table 3 summarises the characteristics of the two real data sets used in our experiments. We note that the University data set corresponds to a larger site, but with a smaller number of links. Also, the Music-Machines data set contains longer sessions on average and with a significantly higher degree of variability. Moreover, while in the University data set 5% of the pages have at least one navigation session starting from them (before HP is tagged onto the session), in the Music-Machines data set the percentage of initial pages is close to 19%.

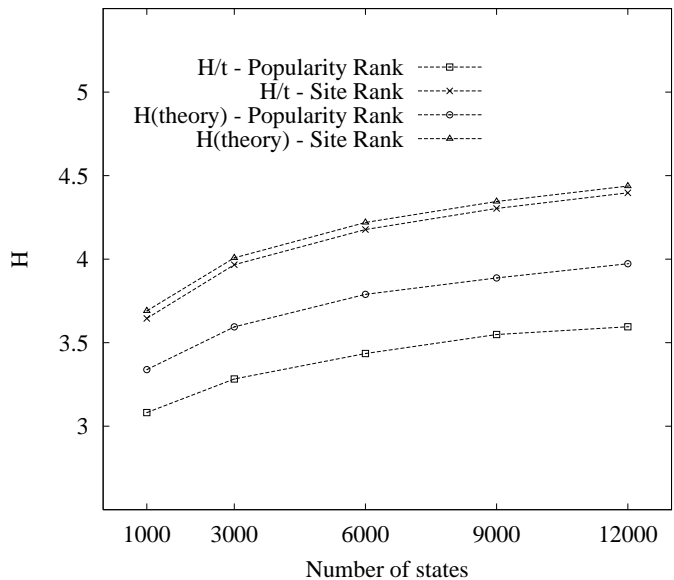

Figure 3: Entropy estimate versus synthetic web site size

| Number of pages | $D^{max}(P\|Q)$ |
|---|---|
| 1000 | 1957.6 |
| 3000 | 5856.3 |
| 6000 | 11674.5 |
| 9000 | 17527.1 |
| 12000 | 23357.5 |

Table 2: Variation of $D^{max}(P\|Q)$ with the number of pages in the web site

Tables 4 and 5 present the results of the runs on the two real data sets, for Popularity Rank and Site Rank, respectively. For each of the data sets we present $H/t$, which is the entropy estimate based on the server log data, $H(\text{theory})$, which is computed using (6), and the running time in seconds of the computation of $H/t$. The Site Rank is estimated by a random walk over the topology generated from the server log data, and its length was set to $10 \cdot (N + L)$ clicks. We note that the estimate for Popularity Rank is very close to the corresponding theoretical value (the difference is only apparent at the sixth decimal place), and, as expected from [LL03], in general, $H/t$ underestimates $H(\text{theory})$. Further, we note that in the real data experiments we cannot take into account unpopular links, since the topology definition is not available.

Table 6 shows the relative entropy between Popularity Rank and Site Rank for the real data sets. As with the artificial data sets the results indicate that the Popularity Rank and Site Rank distributions are very close.

Tables 7 and 8 show the top-ten pages for the Music-Machines site according to Popularity

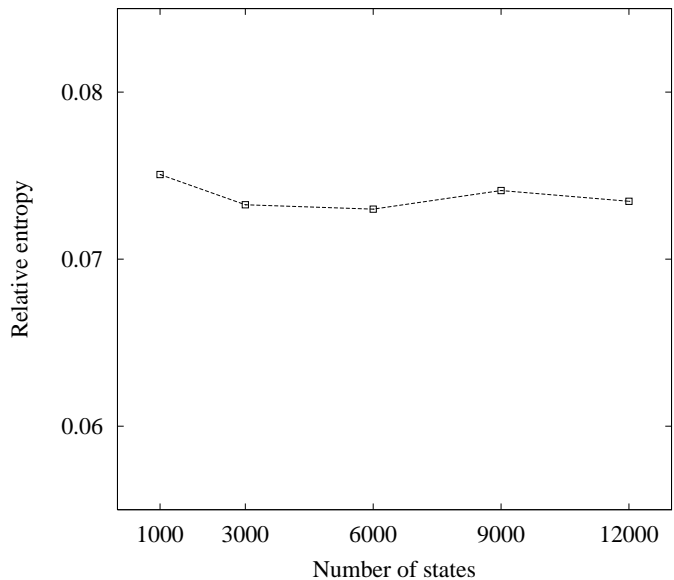

Figure 4: Relative entropy between Popularity Rank and Site Rank versus synthetic web site size

Rank and Site Rank, respectively. As expected, the HP is the most popular page. It is interesting to note that although the top-ten pages according to the two criteria are not in the same ranking order there is only one page in Table 7 that is not in Table 8, which is the page whose rank is 10 in Table 7.

We use a normalised version of Spearman's footrule [FKS03], which is a nonparametric measure that returns a number between 0 and 1, in order to compare the rankings according to Popularity Rank and Site Rank. We subtract 1 from the result to get the complement of Spearman's footrule, since we are interested in the proximity of the rankings rather than in their distance. The measure sums the absolute differences between the ranking or URLs, where a URL that does not appear in a list is ranked at 11, noting that we are only interested in the top-ten pages. For Tables 7 and 8 the value of the complement of Spearman's footrule is 0.7273, indicating a strong correlation between the rankings.

Tables 9 and 10 show the results for the University data set. It is interesting to note that for the University site HP is only the second most popular page according to Site Rank. Also, the degree of overlap between the pages in Tables 9 and 10 is smaller than in the Music-Machines case. For Tables 9 and 10 the value of the complement of Spearman's footrule is 0.6545 also indicating a strong correlation between the rankings but not as strong as the previous case.

## 4 Concluding Remarks

We have investigated two ranking mechanisms for pages in a web site. The first, Site Rank, is an adaptation of PageRank to the granularity of a web site, and the second, Popularity Rank, is based on the frequencies of user clicks on the outlinks in a page that are inferred from server log data and extended to cater for unpopular links. Both the Site Rank and Popularity

| Characteristic | University | Music-Machines |
|---|---|---|
| number of pages | 9,149 | 7,319 |
| number of outlinks | 28,598 | 3,0553 |
| number of sessions | 20,950 | 14,227 |
| number of requests | 106,991 | 104,551 |
| avg. session length | 5.1 | 7.3 |
| stdev. session length | 9.6 | 27.3 |
| max. session length | 396 | 2179 |
| number of initial pages | 479 | 1,376 |
| number of terminating pages | 1,974 | 1,887 |
| avg. number outlinks/page | 3.1 | 4.2 |
| stdev. number outlinks/page | 13.3 | 15.3 |
| avg. number inlinks/page | 2.9 | 4.1 |
| stdev. number inlinks/page | 16.0 | 15.7 |

Table 3: Summary statistics for the real data sets

| Data set | $H/t$ | $t$ | $H$(theory) | time (s) |
|---|---|---|---|---|
| University | 3.30 | 127,942 | 3.30 | 9.69 |
| Music-Machines | 3.94 | 118,779 | 3.94 | 10.23 |

Table 4: Summary of the results for Popularity Rank

Rank can be viewed as the stationary distribution of an ergodic Markov chain modelling user navigation through the web site. For the Site Rank the user is a 'random surfer' and for Popularity Rank the user is navigating according to the frequencies inferred from previous navigation sessions. We have carried out several experiments on artificial and real data sets to compare the distributions of the two rankings mechanisms. Our results show that the relative entropy between the distribution, when normalised, is relatively small: less that 0.1 for the artificial data sets and not much above 0.1 for the real data sets. We conclude from this that the topology of a web site has a strong influence on the navigation patterns of users. This implies, that in practice, Site Rank is a reasonable first order approximation of the aggregate behaviour of users within a web site as given by the Popularity Rank. Moreover, for the real data sets, the top-ten ranked pages are strongly correlated, as indicated by the complement of Spearman's footrule: 0.7273 for the Music-Machines data set and 0.6364 for the University data set.

It is important to note that the result is empirical, and is effected by our choice of normalisation constant, which is the maximum relative entropy. The cases when our conclusion would not be valid can be seen by inspecting (4), in the situation when at many pages being browsed, users consistently choose only one of the links. In practice, in the data sets we have looked at, this did not happen.

We conclude by demonstrating that the distributions of both the Site Rank and the Popularity Rank follow a power law, as was shown in [GP02] for PageRank. Power-law distributions such as these are characteristic of the web [AH01], and have been previously shown for the degree distributions of inlinks and outlinks of the web graph [BKM+00]. We performed linear regression on log-log transformations of the Site Rank and Popularity Rank distributions obtaining power laws with exponents of 1.3180 and 1.4946, respectively, with correlation coefficients of 0.8431 and 0.9294. Log-log plots of frequency against rank for the Music-Machines data set, with the regression line superimposed, are shown in Figures 5 and Figure 6, for the Site Rank and Popularity Rank, respectively. Similarly power law exponents were obtained for the University data set: 1.7604 for Site Rank and 1.4822 for Popularity Rank, with correlation coefficients of 0.9119 and 0.9149, respectively. (We note that we removed the first three points from the Site Rank distribution in both data sets in order to maximise the correlation coefficient.)

| Data set       | $H/t$ | $t$     | $H$(theory) | time (s) |
|----------------|-------|---------|-------------|----------|
| University     | 3.67  | 454,035 | 3.71        | 46.18    |
| Music-Machines | 1.80  | 446,893 | 1.87        | 18.36    |

Table 5: Summary of the results for Site Rank

| Data set       | $D(P\|Q)$ | $D^{max}(P\|Q)$ | Normalised $D(P\|Q)$ |
|----------------|-----------|-----------------|----------------------|
| University     | 681.73    | 5,892.62        | 0.1157               |
| Music-Machines | 960.00    | 7,921.83        | 0.1212               |

Table 6: Summary of the relative entropy

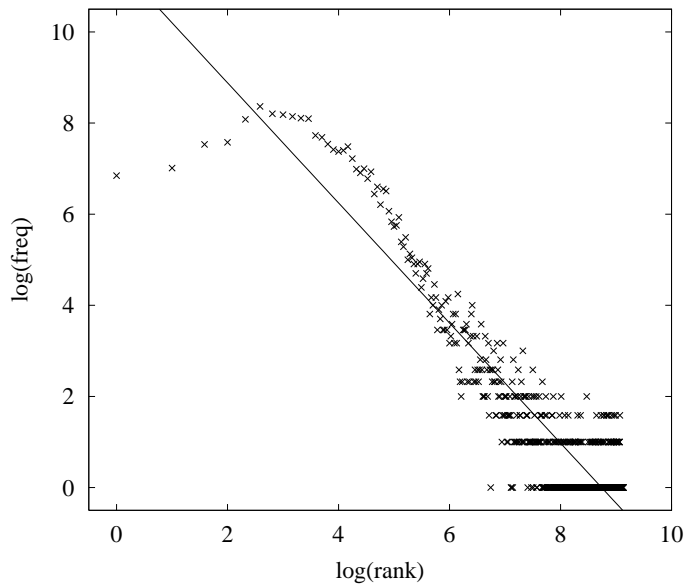

Figure 5: A log-log plot of the Site Rank distribution for the Music-Machines data set

| Rank | URL | Probability |
|---|---|---|
| 1 | HP | 0.1198 |
| 2 | /music/machines | 0.0599 |
| 3 | /music/machines/samples.html | 0.0349 |
| 4 | /music/machines/manufacturers | 0.0308 |
| 5 | /music/machines/samples.html?mmagent | 0.0206 |
| 6 | /music/machines/analogue-heaven | 0.0176 |
| 7 | /music/machines/search.cgi | 0.0147 |
| 8 | /music/machines/manufacturers/roland | 0.0131 |
| 9 | /music/machines/links | 0.0113 |
| 10 | /machines | 0.0110 |

Table 7: Top-ten pages for the Music-Machines data according to Popularity Rank

| Rank | URL | Probability |
|---|---|---|
| 1 | HP | 0.0266 |
| 2 | /music/machines/search.cgi | 0.0120 |
| 3 | /music/machines/manufacturers | 0.0071 |
| 4 | /music/machines | 0.0063 |
| 5 | /music/machines/links | 0.0042 |
| 6 | /music/machines/manufacturers/roland | 0.0027 |
| 7 | /music/machines/analogue-heaven | 0.0027 |
| 8 | /music/machines/guide | 0.0027 |
| 9 | /music/machines/samples.html | 0.0026 |
| 10 | /music/machines/samples.html?mmagent | 0.0026 |

Table 8: Top-ten pages for the Music-Machines data according to Site Rank

| Rank | URL | Probability |
|---|---|---|
| 1 | HP | 0.1638 |
| 2 | /news/default.asp | 0.1533 |
| 3 | /courses/ | 0.0496 |
| 4 | /courses/syllabilist.asp | 0.0280 |
| 5 | /people/ | 0.0215 |
| 6 | /authenticate/login.asp?section=mycti&title=mycti | 0.0190 |
| 7 | /programs/ | 0.0183 |
| 8 | /cti/studentprofile/studentprofile.asp?section=mycti | 0.0183 |
| 9 | /cti/advising/display.asp | 0.0155 |
| 10 | /admissions/ | 0.0113 |

Table 9: Top-ten pages for the University data set according to Popularity Rank

| Rank | URL | Probability |
|---|---|---|
| 1 | /cti/advising/display.asp | 0.0731 |
| 2 | HP | 0.0471 |
| 3 | /news/default.asp | 0.0247 |
| 4 | /courses/ | 0.0104 |
| 5 | /programs/ | 0.0071 |
| 6 | /people/ | 0.0070 |
| 7 | /cti/advising/display.asp?page=coursehistory | 0.0070 |
| 8 | /admissions/ | 0.0068 |
| 9 | /advising/ | 0.0066 |
| 10 | /cti/advising/display.asp?tab=advising&page=communicationlog | 0.0061 |

Table 10: Top-ten pages for the University data set according to Site Rank

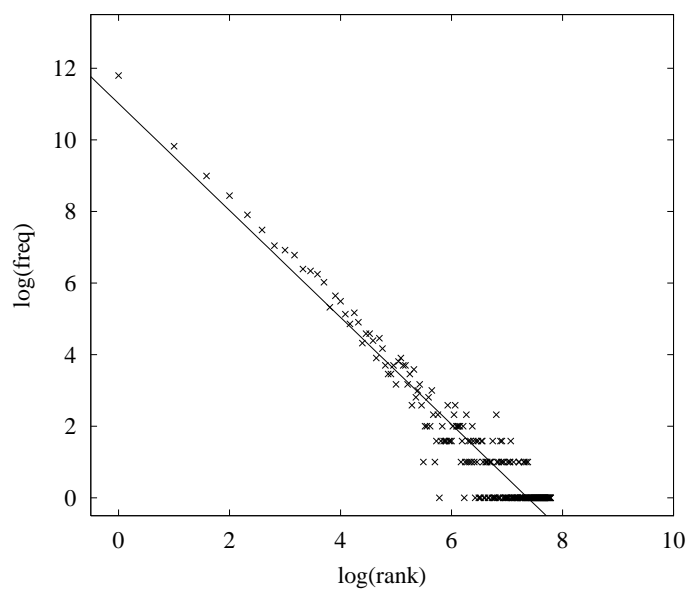

Figure 6: A log-log plot of Popularity Rank distribution for the Music-Machines data set